%% file: main.tex
\documentclass[10pt,twocolumn,letterpaper]{article}

\usepackage{multirow}
\usepackage{graphicx}
\usepackage{wacv}              


\definecolor{wacvblue}{rgb}{0.21,0.49,0.74}
\usepackage[pagebackref,breaklinks,colorlinks,allcolors=wacvblue]{hyperref}

\def\wacvPaperID{*****} 
\def\confName{WACV}
\def\confYear{2026}

\title{Synthesizing Artifact Dataset for Pixel-level Detection}

\author{
Dennis Menn$^1$, Feng Liang$^2$, and Diana Marculescu$^1$\\
$^1$Chandra Family Department of Electrical and Computer Engineering, The University of Texas at Austin\\
$^2$Meta\\
{\tt\small \{dennismenn, dianam\}@utexas.edu, jeffliangf@meta.com}
}

\begin{document}
\maketitle
\input{sec/0_abstract}    
\input{sec/1_intro}
\input{sec/2_related_works}
\input{sec/3_method}
\input{sec/4_settings}
\input{sec/5_experiments}
\input{sec/6_ablation}

\input{sec/7_limitation}

\input{sec/8_conclusion}

\newpage
{
    \small
    \bibliographystyle{ieeenat_fullname}
    \bibliography{main}
}

\end{document}

%% file: sec/0_abstract.tex
\begin{abstract}
Artifact detectors have been shown to enhance the performance of image-generative models by serving as reward models during fine-tuning. These detectors enable the generative model to improve overall output fidelity and aesthetics. However, training the artifact detector requires expensive pixel-level human annotations that specify the artifact regions. The lack of annotated data limits the performance of the artifact detector. A naive pseudo-labeling approach-training a weak detector and using it to annotate unlabeled images-suffers from noisy labels, resulting in poor performance. 
To address this, we propose an artifact corruption pipeline that automatically injects artifacts into clean, high-quality synthetic images on a predetermined region, thereby producing pixel-level annotations without manual labeling.
The proposed method enables training of an artifact detector that achieves performance improvements of 13.2\% for ConvNeXt and 3.7\% for Swin-T, as verified on human-labeled data, compared to baseline approaches. This work represents an initial step toward scalable pixel-level artifact annotation datasets that integrate world knowledge into artifact detection.
\end{abstract}

%% file: sec/1_intro.tex
\section{Introduction}
\begin{figure}
    \centering
    \includegraphics[width=0.48\textwidth]{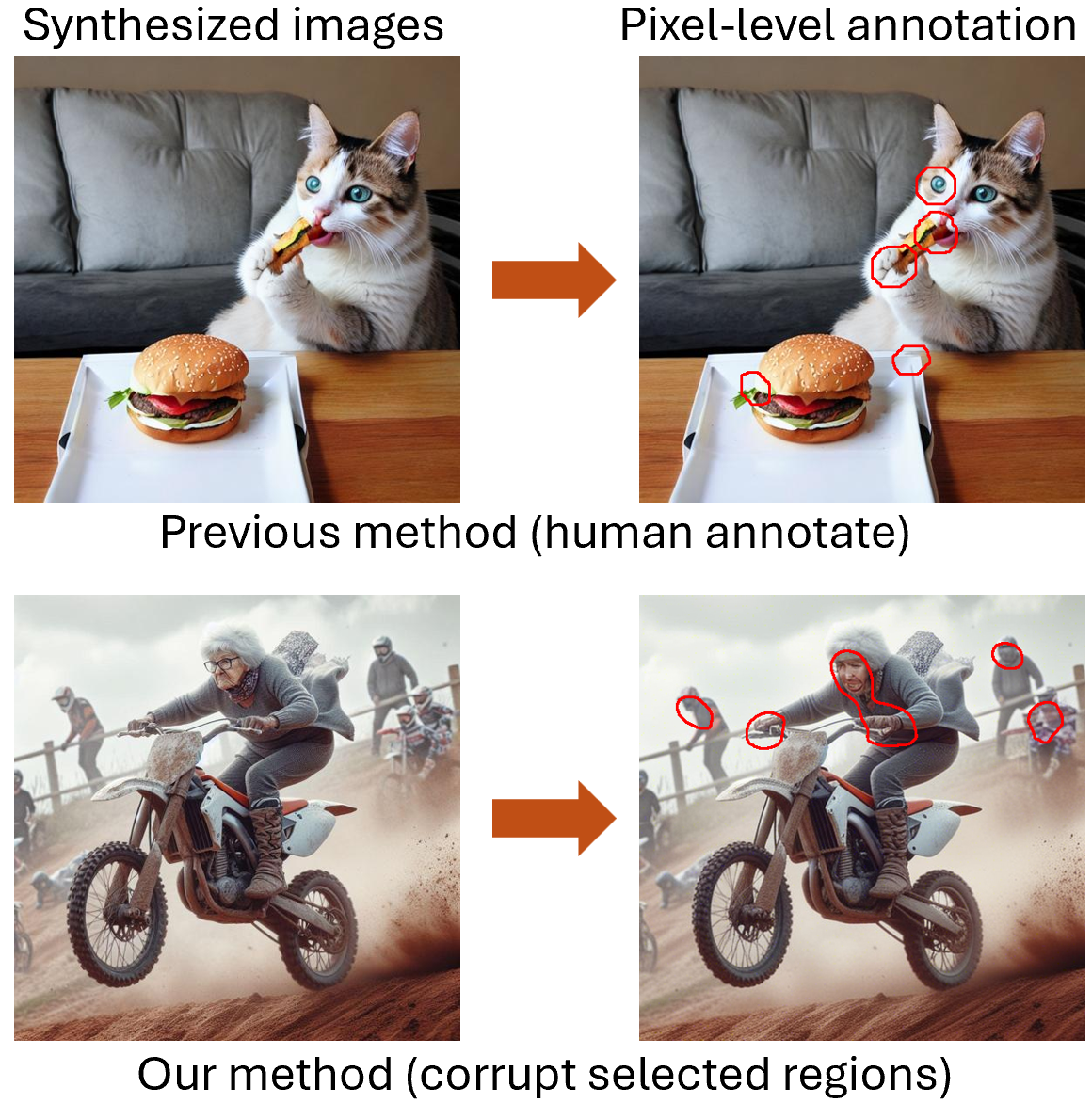}
    \caption{\textbf{Overview of our proposed method:} Unlike previous methods that rely on human annotation to identify artifact regions, our proposed approach introduces artifacts directly into selected regions to form a pixel-level annotation dataset.}
    \label{fig:paper_overview}
\end{figure}

Recent advancements in generative models for images and videos have led to remarkable progress in producing high-quality synthetic content \cite{ho2020ddpm, saharia2022imagine, Rombach2022ldm}. However, these models often generate artifacts, particularly in complex scenes involving intricate details like human faces, fingers, or textured objects \cite{richhf2024liang, zhang2022perceptual, menn2025similarity}. These artifacts appear as distortions, unnatural blending, or structural inconsistencies, compromising the realism of the generated outputs. Despite significant computational resources for training and improving generative models, mitigating these artifacts remains a persistent challenge \cite{xing2025focus, richhf2024liang}.

Artifact detectors are models used to identify and localize artifacts at the pixel level in synthesized images. Recent studies on artifact detectors provide a way for guiding generative models to reduce artifacts produced by the generative model, enhancing the appeal of the generated content to humans \cite{Xu2023ImageReward, wallace2023diffusion, xing2025focus}. The methods include their use as reward models for finetuning the generative model \cite{wu2023alsd, kirstain2023pickapic, Xu2023ImageReward, wallace2023diffusion, xing2025focus}. However, the scarcity of pixel-level annotations limits the performance of artifact detectors. This is because human annotation of artifacts is costly, labor-intensive, and complicated by the variability in artifact appearance, which ranges from subtle distortions to prominent structural flaws \cite{richhf2024liang, menn2025similarity}. For instance, \citet{richhf2024liang} reports that annotating each image takes approximately ten minutes; however, their process includes not only artifact annotation but also identifying misalignment and rating overall image quality, typically involving an average of three annotators. In contrast, \citet{zhang2022perceptual} notes that artifact annotations take about one minute per image. To ensure reliability of the annotation, multiple annotators are often required to annotate per image.

In this paper, we propose a novel framework for automatically generating pixel-level artifact annotations to enhance the artifact detector's performance in a weakly-supervised setting. Unlike previous methods that rely on human evaluators to annotate artifact locations, we utilize an artifact corruption pipeline to generate artifacts in high-quality synthetic images for the selected regions, as shown in Figure \ref{fig:paper_overview}. By employing latent inversion techniques \cite{garibi2024renoise}, we manipulate the latent representations of images to simulate realistic artifacts produced by generative models. This pipeline allows artifacts to be confined to the selected regions, which are prone to artifacts in real-world scenarios, while preserving the image content of non-selected areas. This allows for creating a pixel-level artifact dataset by combining the selected region for corruption as pixel-level annotations with the corrupted image to form the dataset. We summarize the contributions of this paper as follows:
\begin{itemize}
    \item We formulate a novel framework for automatically generating pixel-level artifact datasets.
    \item We introduce a method for confining synthetic artifacts to specific image regions and enhancing their quality so they are close to artifacts generated in the wild by existing generative models.
    \item We demonstrate that artifact detectors trained on our synthetic dataset improve performance in a weakly supervised setting, outperforming direct training with human label samples.
\end{itemize}

This framework can be viewed as a first step towards scalable pixel-level annotation for training artifact detectors, enabling the incorporation of world knowledge by specifying object inconsistency to the model.

%% file: sec/2_related_works.tex
\section{Related Work}
\subsection{Artifact Detection}
Artifact detection is critical in image/video generation research due to the frequent occurrence of undesirable artifacts that degrade image quality \cite{wu2023alsd, Xu2023ImageReward, zhang2022perceptual, richhf2024liang}. Research in artifact detection aims to automatically identify artifacts and enhance generative models to minimize their occurrence \citep{rafailov2023direct, xing2025focus}. A key first step in developing artifact detectors is creating datasets, which are divided into two types: (1) holistic human preference datasets and (2) region-specific datasets. Holistic datasets rely on user-preferred generated images to train models that select images based on overall human appeal \cite{wu2023alsd, wu2023human2, kirstain2023pickapic, Xu2023ImageReward}. In contrast, region-specific datasets require labor-intensive pixel-level annotations to pinpoint artifact locations \cite{zhang2022perceptual, richhf2024liang}. Unlike prior work, our approach focuses on automatically creating pixel-level annotations for artifact detection.

Artifact detectors play a crucial role in enhancing generative model performance by identifying and mitigating artifacts generated \citep{zhang2022perceptual, richhf2024liang, wu2023alsd, Xu2023ImageReward, wallace2023diffusion, xing2025focus}. \citet{zhang2022perceptual} uses detectors to guide inpainting of artifact regions, improving image quality. Similarly, \citet{richhf2024liang} shows that fine-tuning models with images tailored by artifact detectors boosts output quality. Several studies further demonstrate that optimizing diffusion models with artifact detectors or human preference-based reward models - through techniques like reward feedback learning \cite{Xu2023ImageReward} or Direct Preference Optimization \cite{rafailov2023direct} - enhances generative model performance \cite{wu2023alsd, Xu2023ImageReward, wallace2023diffusion, xing2025focus}, highlighting the pivotal role of artifact detection in generative modeling.

\subsection{Mechanisms Behind Artifacts Formation}
Understanding the cause of artifacts is essential for generating realistic ones. While the precise mechanisms remain elusive, several studies offer insights. \citet{menn2025similarity} observed that artifacts occur when denoised images from consecutive diffusion steps differ significantly, suggesting that artifacts likely result from overlapping distinct objects in the generated content. \ \citet{aithal2024understanding} argue that artifacts occur when generated data lies between modes of the true data distribution. Meanwhile, \citet{aithal2024understanding, hong2022improving} propose that self-attention mechanisms in early denoising steps may amplify or suppress features, leading to hallucinations. Despite these findings, the definitive indicator of artifact formation remains elusive, highlighting the need for further exploration and elucidating difficulties in generating realistic artifacts.

\subsection{Image Editing Techniques}
Image editing techniques allow photorealistic modifications to specific regions of an image - usually guided by text prompts - while keeping the rest unchanged. Denoising Diffusion Implicit Models (DDIM) \cite{song2021denoising} enable such editing by gradually adding noise and then modifying the image by adding noise iteratively during the forward diffusion process. However, DDIM inversion can result in divergence from the original image due to nonlinearities and imprecise score estimates. ReNoise \cite{garibi2024renoise} addresses this by iteratively refining predictions with a pretrained diffusion model. Mask-guided image editing methods such as \citet{Avrahami_2022_blended, couairon2023diffedit, wang2023instructedit} further improve control by restricting edits to masked regions, ensuring unmasked areas remain unchanged. Our work builds on these approaches by introducing artifacts only in masked regions while preserving the rest of the image.

%% file: sec/3_method.tex
\section{Methods}
\label{sec:method}
\begin{figure*}[t]
  \centering
  \includegraphics[width=0.8\textwidth]{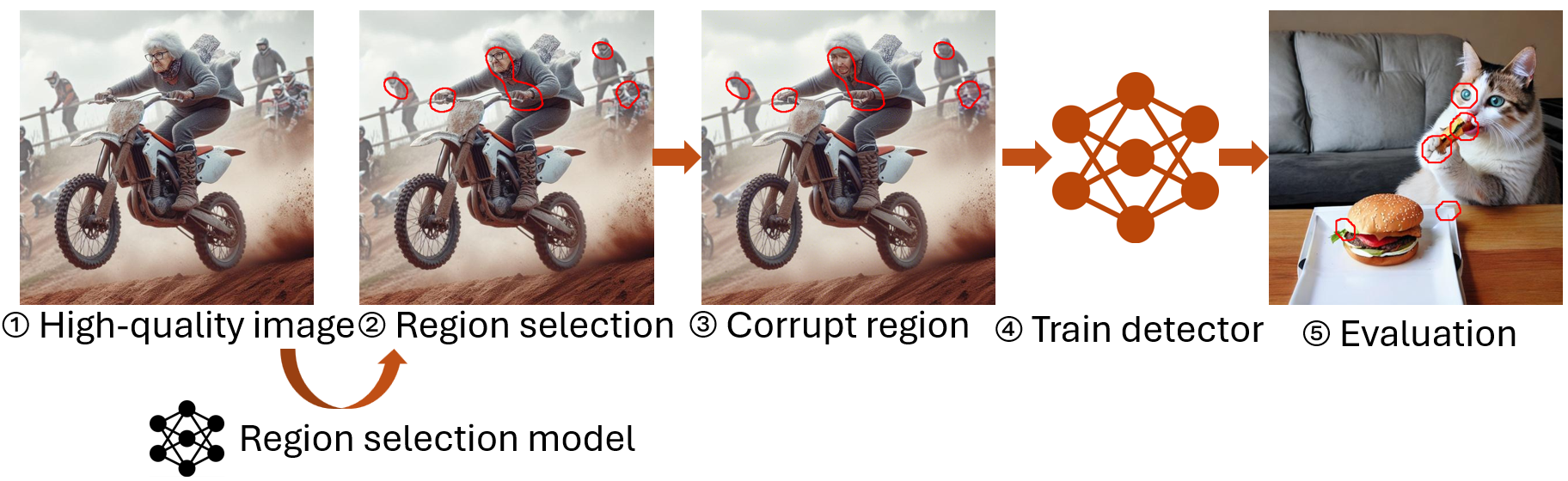}
  \caption{\textbf{Overview of the artifact dataset creation pipeline:} We first select high-quality synthesized images. We then identify regions to corrupt using a region selection model. Next, we introduce artifacts into the selected regions via the corruption pipeline. These artifact images, paired with their selected regions, form a pixel-level annotated artifact dataset used to train the artifact detector. The detector is then evaluated on a human-annotated dataset.}
  \label{fig:method_overview}
\end{figure*}

The following properties for synthesized artifacts \( A_s \) is needed to form a pixel-level artifact dataset:
\begin{enumerate}
    \item \textbf{Artifacts region confinement:} Artifacts \( A_s \in \mathbb{R}^{H \times W} \) are confined to selected regions defined by the predetermined binary mask \( M_s \in \mathbb{R}^{H \times W} \), where \( M_s[i,j] \in \{0, 1\} \). Specifically,
    \begin{equation}
    A_s = A_s \odot M_s.
    \end{equation}
    ensuring artifacts only appear where \( M_s = 1 \).
    \item \textbf{Appearance}: The visual characteristics of \( A_s \) closely resemble those of wild artifacts \( A_w \) observed in synthesized images, denoted as \( A_s \sim A_w \).
\end{enumerate}
Then, the dataset, consisting of binary masks and corrupted images, is used to train an artifact detector. 

\noindent \textbf{Region Selection Model}:
Artifacts in synthesized images \( I \in \mathbb{R}^{H \times W \times C} \) exhibit a non-uniform spatial distribution, concentrated in regions \( M_w \in \mathbb{R}^{{H \times W}} \), where $M_w[i,j] \in \{0, 1\}$ , such as faces or fingers~\cite{richhf2024liang}. To replicate this in synthetic artifacts, we train a region selection model $f_\theta(\cdot)$ to predict artifact-prone regions \( M_s \). The model, initialized with a pre-trained object segmentation model, is finetuned on a dataset \( \mathcal{D}_{human} = \{ (I_i, M_i) \}_{i=1}^N \), where \( M_i : \mathbb{R}^{{H \times W}} \to \{0, 1\} \) denotes human-annotated artifact masks (labels). The objective is to minimize a segmentation loss, ensuring that the selected regions \( M_s = \{ x \in \mathbb{R}^{H \times W} \mid f_\theta(I_i)(x) \geq \tau \} \) induce a synthetic artifact distribution \( p_s(x \mid I) \), representing the likelihood of regions containing artifacts, that closely aligns with the real-world artifact distribution \( p_w(x \mid I) \), \textit{i.e.}, \( p_s \approx p_w \).

\subsection{Artifact Dataset Creation Pipeline}
In Figure \ref{fig:method_overview}, we show the process of creating a pixel-level artifact dataset and evaluating its quality by training an artifact detector on top of it. In the following, we introduce each component of the pipeline.
\begin{enumerate}
    \item \textbf{Image Selection}: We use a dataset of high-quality synthetic images with few pre-existing artifacts. This helps ensures that areas outside the artifact-introduction area feature few artifacts, and the selected region exhibits a high concentration of artifacts after corruption, offering higher-quality pixel-level annotations.    
    \item \textbf{Region Selection}: Given an high-quality image $I$, the trained region selection model identifies artifact-prone regions $M_s = \{ x \in \mathbb{R}^{{H \times W}} \mid f_\theta(I)(x) \geq \tau \}$ (e.g., faces, fingers) within each image. These selected regions, are where we introduce artifacts.
    \item \textbf{Artifact Corruption}: A corrupted image $I^{corr.}$ is created via applying corruption pipeline toward the selected region on latent $z$ in different time step, \textit{i.e.} $I^{corr.}=\text{Corr.}(M_s, z_t, z_0)$, such that the newly introduced artifacts $A_s \sim A_w$ that mimic the characteristics of artifacts generated in the wild, while the rest of the image remains mostly unaltered.
    \item \textbf{Training the Artifact Detector}: The corrupted images, paired with their corresponding selected region, $\mathcal{D}_s = \{ (I^{corr.}_i, M_s^i) \}_{i=1}^N$ form a pixel-level artifact dataset. This dataset is then used to train an off-the-shelf artifact detector $D_\phi(\cdot)$.

     \item \textbf{Evaluation of Artifact Dataset Quality}:
To assess the quality of the synthetic artifact dataset, we evaluate the trained artifact detector on a test set comprising images and annotations pairs $(I^{test}_i, M_i) \in \mathcal{D}_{test} $ with artifacts generated in the wild and corresponding human-annotated, pixel-level annotations. The alignment between the detector’s predictions and human annotations $D_{\phi}(I_i^{test}) \sim M_i$ is quantified to determine the ability of synthetic artifacts to replicate human-perceived artifacts. 
\end{enumerate}

\begin{figure*}[t!]
  \centering
   \includegraphics[width=0.8\linewidth]{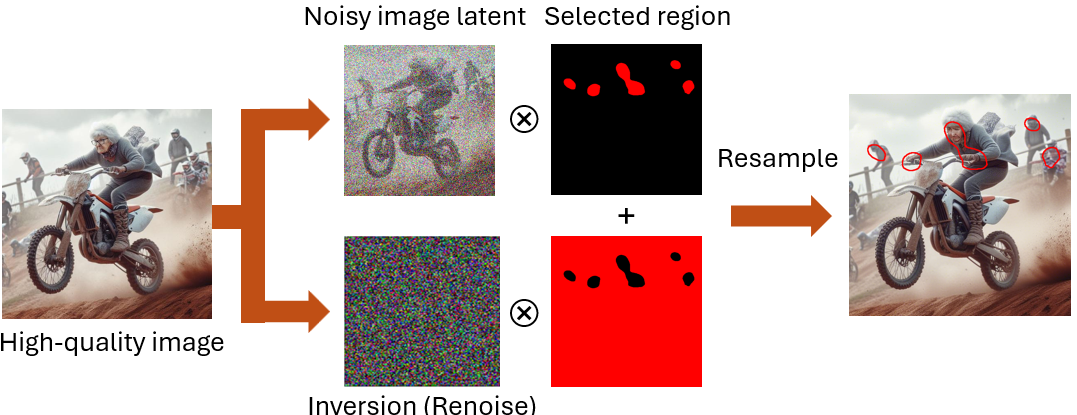}
   \caption{\textbf{The Artifact Corruption Pipeline:} Artifacts are introduced to a selected image region by first performing latent inversion on the entire image using the ReNoise algorithm \cite{garibi2024renoise}. The selected region’s latent representation is then replaced with the image latent with the addition of Gaussian noise. The modified latent is then resampled to generate the final image. Empirical results show that this process can induce artifacts in the selected regions.}
   \label{fig:corruption_pipeline}
\end{figure*}

\subsection{Artifact Corruption Pipeline}
\label{sec:corruption_method}
The artifact corruption pipeline aims to introduce synthetic artifacts into selected regions of high-quality images $I$ while preserving the content of the remaining areas. As shown in Figure \ref{fig:corruption_pipeline}, there are three steps for the corruption pipeline: latent inversion, region-specific corruption, and resampling. In the following, we detail each component.

\paragraph{Latent Inversion:} To generate realistic artifacts, we first perform latent inversion on the initial image latent \( z_0 \) to obtain \( z_t \). Direct corruption of the image latent often results in simplistic corruptions, failing to capture the diverse variability of artifacts. In contrast, the resampling process following latent inversion enhances the diversity of corruptions by enabling the diffusion model to adapt them to the surrounding context. Moreover, latent inversion techniques effectively preserve the content of non-selected regions, ensuring artifacts are confined to selected areas \cite{garibi2024renoise}, thereby making this approach suitable for generating artifacts. The inversion follows the recursive equation:

\begin{equation}
z_t = \frac{z_{t-1} - \psi_t \epsilon_\theta(z_t, t, c) - p_t \epsilon_t}{\phi_t}.
\end{equation}
where \( \epsilon_\theta \) denotes the denoiser output and \( \epsilon_t \) represents sampled noise.

\paragraph{Region Specific Corruption:} To localize artifacts in selected regions, we process the inverted latents of selected (\( z^{\text{sel}}_t \)) and non-selected (\( z^{\text{non}}_t \)) regions differently. For non-selected regions, we preserve the original content by keeping the latent unchanged after inversion:
\begin{equation}
z_t^{\text{non}} = z_t.
\label{eqn:non}
\end{equation}
For selected regions, we introduce artifacts by adding Gaussian noise \( \epsilon_t \) with magnitude \( p_t' \) and adjusting image latent \( z_0 \) to match the inversion steps $t$ of non-selected regions:
\begin{equation}
z_t^{\text{sel}} = \phi_t' z_0 + p_t' \epsilon_t.
\label{eqn:sel}
\end{equation}
Prior to combining latents from selected and non-selected regions. Using masks \( \mathbf{1} - M_s \) and \( M_s \), where \( \mathbf{1} \in \mathbb{R}^{H\times W} \), to designate selected and non-selected regions is needed:
\begin{equation}
\text{Corr.}(M_s, z_t, z_0) = \lvert \mathbf{1} - M_s \rvert \otimes z_t + M_s \otimes (\phi z_0 + p \epsilon).
\end{equation}

The corruption function enables spatially targeted synthetic artifacts, mimicking artifacts in the wild after resampling. Empirically, this approach generates artifacts in selected regions due to (1) the inherent difficulty of generating regions like faces or fingers, and (2) out-of-distribution effects from discrepancies between noise-corrupted selected latents and intact non-selected latents from inversion. 

\paragraph{Resampling:} Finally, we resample \( z_t^{\text{corr.}} \) back to time step \( t=0 \) using the recursive sampling algorithm:
\begin{equation}
z_{t-1}^{\text{corr.}} = \phi_t z_t + \psi_t \epsilon_\theta(z_t, t, c) + p_t \epsilon_t.
\end{equation}
This will lead to a range of artifacts/distorted objects that mirror the artifacts in the wild.

%% file: sec/4_settings.tex
\section{Experimental Setup}
\label{sec:exp_setting}
This section outlines the experimental setup, including the selection of datasets, training and evaluation pipeline for the artifact detector, and the hyperparameters used for the corruption pipeline.

\subsection{Dataset Selection}
In our experiment, we utilize the RichHF-18k Dataset \cite{richhf2024liang} and the Human Preference Synthetic Dataset (HPS Dataset) \cite{Egan_HPS_2024}. 

\paragraph{RichHF-18k:}
The RichHF-18k dataset offers human-annotated, pixel-level artifact labels, making it ideal for training the region selection model and evaluating the artifact detector. However, the label values are continuous. Therefore, we preprocess these continuous artifact probability labels into binary labels by excluding those labels that have fewer than three annotators. Annotations are classified as indicating the presence of artifacts only if at least two out of three human annotators indicate the presence of artifacts. This preprocessing results in 7,801 training samples and 500 test samples with binary labels.
To train the region selection model, we randomly select 100 images from the training set, allowing it to identify regions that are prone to artifacts. We also utilize images from the RichHF training set as a source for the corruption pipeline. For evaluating the performance of the artifact detector, we use the test set from RichHF-18k.


\paragraph{HPS Dataset:}
The HPS Dataset is a high-quality synthetic dataset containing over one million images, primarily generated by DALL-E 3, selected and posted by human users \cite{Egan_HPS_2024}. We randomly choose 10,000 images from HPS Dataset for the corruption pipeline. These images are chosen to minimize pre-existing artifacts, ensuring that artifacts do not exist outside the selected regions for corruption. These images are then sent to the corruption pipeline and used for training the detector.

\subsection{Region Selection Model Training}
Artifacts in images often concentrate on specific regions. To replicate this distribution, we train a region selection model to identify artifact-prone regions rather than corrupting random areas. The region selection model is finetuned on an object segmentation model (Swin-T or ConvNeXt, as specified in the experiments), pre-trained on the ADE20K dataset \cite{zhou2017scene} using 100 human-annotated samples from the RichHF-18k training set. We trained the model for 1,000 iterations with a batch size of eight, achieving convergence before completing all training iterations. 

\subsection{Artifact Corruption Pipeline Parameters}
The artifact corruption pipeline is configured with the following hyperparameters to ensure artifacts are introduced in the selected region and realistic artifact generation:

\begin{enumerate}[label=$\bullet$]
    \item \textbf{Inversion Algorithm}: We employ the ReNoise algorithm for latent inversion \cite{garibi2024renoise}, which is known for preserving image content in non-selected regions.
    \item \textbf{Sampling Algorithm}: The DDIM-50 algorithm \cite{song2021denoising} resamples the combined latent.
    \item \textbf{Inversion Steps}: The latent is inverted for 10 steps to balance content preservation and artifact induction.
    \item \textbf{Noise Addition}: For the selected regions, Gaussian noise is added at a level corresponding to the last 10 steps in the DDIM-50 schedule, aligning with the inversion algorithm.
\end{enumerate}

\subsection{Training and Evaluating the Artifact Detector}
The artifact detector is a segmentation-based model, with the same architecture as the region selection model, trained on samples from the HPS or RichHF Dataset to predict pixel-level artifact locations. Training is conducted for 10,000 iterations, achieving convergence within the training iterations. We use the AdamW optimizer with a learning rate of $6 \times 10^{-5}$, betas of (0.9, 0.999), and a weight decay of 0.01. The training objective is minimized using cross-entropy loss, ensuring accurate segmentation of artifact regions. Completing each experiment on an eight A5000 GPU cluster takes around four hours. 

After training, the artifact detector is evaluated on the RichHF-18k test set. We compute standard segmentation metrics using Intersection over Union (IoU) to assess the alignment between the detector’s predictions and human annotations. This evaluation quantifies the detector’s ability to identify artifacts in a way consistent with human perception, hence validating the quality of the synthetic artifact dataset.

%% file: sec/5_experiments.tex
\section{Experiments}
\label{sec:exp}

\begin{figure*}[tbp]

\hspace{-2em}

   \includegraphics[width=1.1\linewidth]{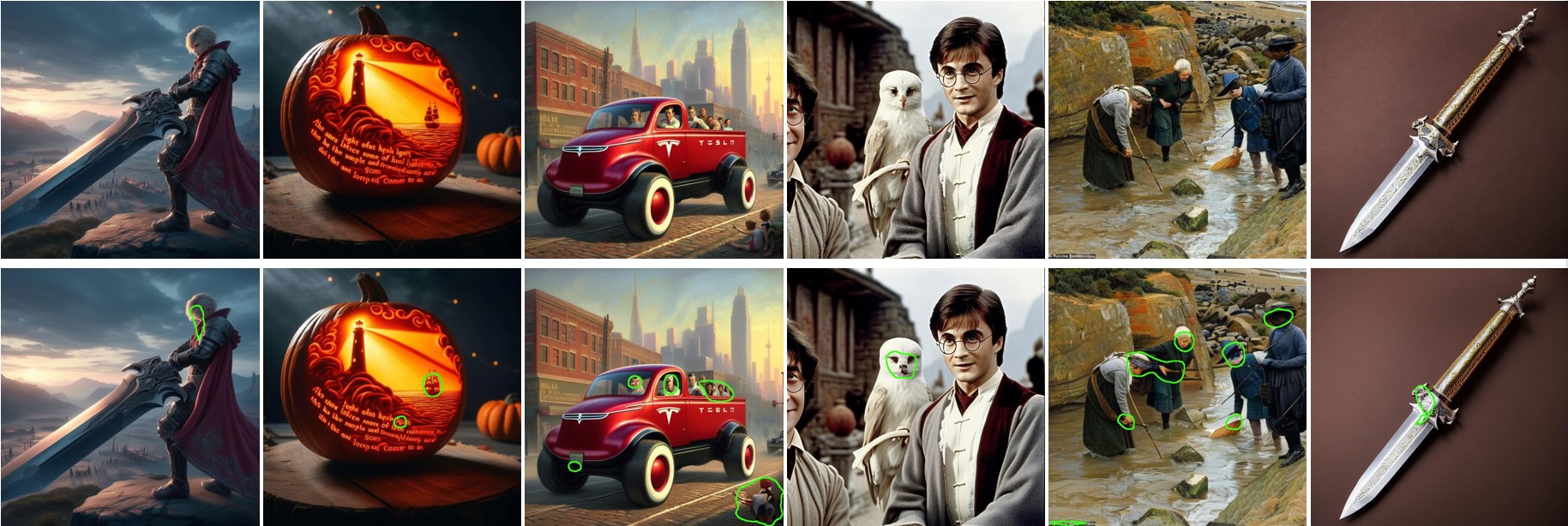}
   \caption{\textbf{Introducing artifacts to images.} The top half row displays original images, while the bottom half row shows corrupted images with green lines denoting the selected region for artifacts.} 
   \label{fig:experiment_demo}
\end{figure*}
\subsection{Qualitative Evaluation of Artifact Corruption}
Figure \ref{fig:experiment_demo} shows corrupted images - the left three from the HPS dataset and the right three from RichHF - processed by our artifact corruption pipeline. The upper half of each figure displays the original images. The lower half displays the corrupted versions. Green circles indicate the regions selected for artifact introduction, which subsequently serve as pixel-level annotations for training. The figure illustrates the proposed method's ability to generate artifacts within these selected regions. For example, in the three images on the left of Figure \ref{fig:experiment_demo}, the swordsman's face shows a noticeable distortion; the boat and the text on the lantern are altered. Similarly, the two children in the lower right corner appear merged, and the driver’s facial expression is distorted. A detailed examination of the figure reveals these changes more clearly. Despite being able to introduce artifacts in the region, the pipeline is not perfect. Not all selected regions produce artifacts; for example, the swordsman’s collar in the first image from the left in Figure \ref{fig:experiment_demo} remains unaffected. Additionally, artifacts may occasionally pre-exist outside the selected region, as observed with distorted text in the second image on the left from Figure \ref{fig:experiment_demo}. However, we observe that the selected regions are notably more likely to contain artifacts post-corruption. 

\subsection{Quantitative Evaluation of Artifact Detector}
Table \ref{tab:exp_results} presents the effectiveness of the proposed corruption pipeline in enhancing artifact detector performance across different datasets and model architectures with limited ground-truth labels. Detector performance is evaluated on the RichHF-18k test set with human annotations and measured by the mIoU score. 

\begin{table}[htbp]
\centering
\caption{\textbf{Artifact detector performance comparison}: our model versus baselines on RichHF-18k and HPS datasets, using two model architectures, with mIoU scores ($\%$) evaluated on the RichHF test set.}
\label{tab:exp_results}
\resizebox{\columnwidth}{!}{
\begin{tabular}{clcc}
\toprule
\textbf{Train Set} & \textbf{Config.} & \textbf{ConvNeXt} & \textbf{Swing-T}  \\
\midrule
RichHF-100 & Human labels & 23.27 & 26.23 \\
\midrule
\multirow{2}{*}{HPS} & Pseudo labels & 23.15 & 26.51 \\
                     & \textbf{Pseudo labels + Corr.} & \textbf{26.35} & \textbf{27.30} \\
\midrule
\multirow{2}{*}{RichHF-7701} & Pseudo labels & 24.32 & 27.36 \\
                             & \textbf{Pseudo labels + Corr.} & \textbf{26.52} & \textbf{28.09} \\
\bottomrule
\end{tabular}
}
\end{table}

Columns labeled “RichHF-100” refer to an artifact detector trained exclusively on 100 human-annotated samples from the RichHF-18k trainset. The column "Pseudo labels" denotes a detector trained on in-the-wild images from either the HPS or RichHF datasets. In this case, pseudo-labels for artifact regions are generated by the region selection model, which is the RichHF-100 model. The “Pseudo labels + Corr.” setting applies the corruption pipeline to the “Pseudo labels” configuration. This process introduces artifacts into regions based on pseudo-labels using the proposed corruption pipeline.

Given the same human-labeled data, Table \ref{tab:exp_results} shows that the corruption pipeline enhances detector performance compared to the baselines across various model architectures and datasets. For the HPS data set, the corruption pipeline increases mIoU by 13.2\% for ConvNeXt and 3.1\% for Swin-T compared to directly training the model with human-annotated data. The method also outperforms the pseudo label setting by 13.8\% for ConvNeXt and by 2.9\% for Swin-T. This demonstrates the effectiveness of the corruption pipeline, as the only difference between the two scenarios is the presence or absence of the corruption pipeline.

Similar improvements are observed across both RichHF and HPS datasets and for both Swin-T and ConvNeXt, indicating the generalizability of the corruption pipeline in enhancing artifact detector performance.
Notably, detectors trained on RichHF-18k outperformed those trained on the HPS dataset. This difference likely stems from a domain shift, as the RichHF-18k test set aligns more closely with its training distribution.

\begin{figure*}[htb]
\centering
\begin{minipage}{0.75\textwidth}
  \centering
  \includegraphics[width=1\linewidth]{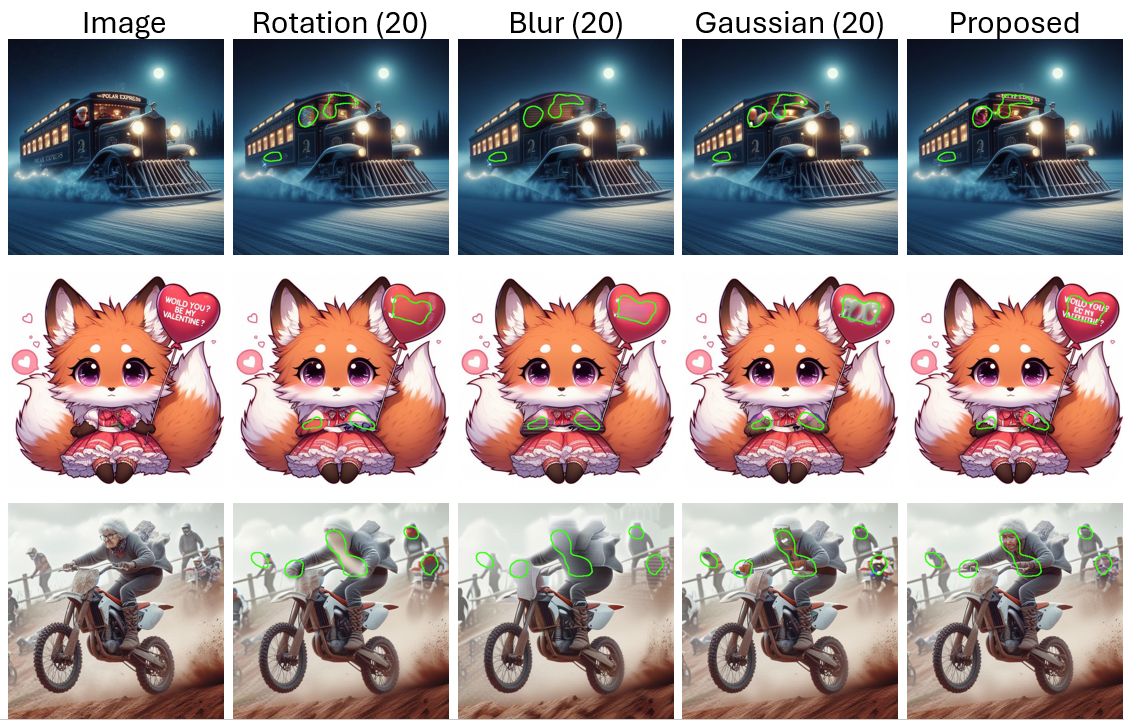}
  \caption{\textbf{Visualization of different corruption methods.} Different corruption methods produce different effects on selected image regions.} 
  \label{fig:diff_corr}
\end{minipage}
\hfill
\begin{minipage}{0.21\textwidth}
  \centering
  \vspace{2.5 em}
  \includegraphics[width=0.7\linewidth]{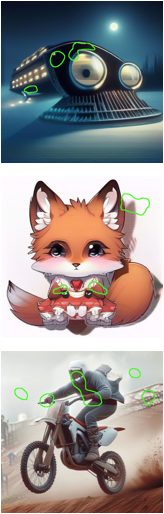}
  \caption{\textbf{Visualization on corrupt images in the initial latent} (invert 50 steps).}
  \label{fig:invert_init}
\end{minipage}
\end{figure*}

\subsection{Comparison with Other Corruption Methods}
We evaluated various corruption methods to generate artifacts in the selected regions, aiming to identify a reliable technique that meets two key constraints: (1) confinement of artifact regions to the selected region and (2) the appearance of synthesized artifacts that closely resemble real artifacts, as mentioned in Sec. \ref{sec:method}. The corruption methods we investigated include blurring, downscaling the latent resolution by \(8\times\), rotating the image by 90 degrees, directly replacing the latent with Gaussian noise, and our proposed method that integrates Gaussian noise into the image latent based on time steps. We further corrupt and resample the latent for 20 steps, except for the proposed method, to better blend the corruptions with the surroundings. The corrupted results are shown in Figure \ref{fig:diff_corr}, indicating that rotation and blur often cause objects in the selected region to vanish. However, such an object's disappearance does not align with artifacts generated in the wild. When directly changing the inverted image latent with Gaussian noise $N(0,1)$, resampling will adapt the noise with its surroundings, but can lead to weird objects, such as the words on the fox's balloon appearing as a gray object due to a mismatch between the noise magnitude and the sampling time step.

We compare these methods by training an artifact detector on the corrupted images and evaluating it on the human-annotated RichHF-18k test set. As shown in Table \ref{tab:diff_corr}, the proposed method achieved the highest mIoU score, indicating a better performance in generating realistic artifacts. Blur performed the worst, with a mIoU score below four, well below the baseline from training the model with human labels. This happened because the model misidentifies blurred areas as artifacts. The experiment highlighted that effective latent space corruption methods are essential for improving the artifact detector; otherwise, the artifact introduction method may mislead it.

\begin{table}[htbp]
\centering
\caption{\textbf{Quantitative analysis of different corruption methods}. mIoU scores (\%) are evaluated on the
RichHF test set.}
\label{tab:diff_corr}
\resizebox{0.7\columnwidth}{!}{
\begin{tabular}{lcc}
\toprule
Corr. Method & ConvNext & Swing-T \\
\midrule
Baseline & 23.27 & 26.23 \\
\midrule
Rotation & 14.93 & 17.40 \\
Blur & 3.03 & 1.73 \\
Gaussian & 12.52 & 13.86 \\
\textbf{Proposed} & \textbf{26.35} & \textbf{27.30} \\
\bottomrule
\end{tabular}
}
\end{table}

\begin{table}[htb]
\centering
\caption{\textbf{Quantitative analysis for corrupting image latents at different time steps}. Corruption between steps 40–45 yields higher mIoU, and the corruption pipeline from different time steps consistently outperforms baseline methods. }
\label{tab:diff_timesteps}
\resizebox{\columnwidth}{!}{
\begin{tabular}{lcccc}
\toprule
\multirow{2}{*}{\textbf{Configuration}} & \multicolumn{2}{c}{\textbf{RICHHF-18k}} & \multicolumn{2}{c}{\textbf{PROGAMER}} \\
\cmidrule(lr){2-3} \cmidrule(lr){4-5}
& \textbf{Swing-T} & \textbf{ConvNext} & \textbf{Swing-T} & \textbf{ConvNext} \\
\midrule
RichHF-100 & 26.23 & 23.27 & 26.23 & \textbf{23.27} \\
Pseudo labels & \textbf{27.36} & \textbf{24.32} & \textbf{26.51} & 23.15 \\
\midrule
Corr. (45) & \textbf{28.48} & \textbf{26.71} & 27.16 & 26.24 \\
Corr. (40) & 28.09 & 26.52 & \textbf{27.30} & \textbf{26.35} \\
Corr. (35) & 27.70 & 25.63 & 27.03 & 26.17 \\
Corr. (30) & 27.94 & 24.87 & 26.44 & 25.91 \\
\bottomrule
\end{tabular}
}
\end{table}

%% file: sec/6_ablation.tex
\section{Ablation Study}
In this section, we evaluate designing choices for generating realistic artifacts, focusing on corruption time step inversion strategies for latent corruption, and exploring the importance of resampling the corrupted latents.

\subsection{Selecting Corruption Time Steps}
We investigated the influence of corruption at different time steps on artifact generation using the method proposed in Section \ref{sec:corruption_method}. We found that corruption at time step zero often alters the entire image after resampling 50 steps, failing to restrict artifacts to the selected region, and resulting in poor pixel-level annotations, as shown in Figure \ref{fig:invert_init}. The train's shape is completely distorted, rendering the selected region an unreliable indicator of artifacts. 

In Table \ref{tab:diff_timesteps}, we train the artifact detector with images corrupted on different time steps, which reveal that corruption occurring between time steps 45 and 40 consistently produces higher mIoU. It is important to note that, regardless of the time step used for corruption, the trained artifact detector consistently outperforms those trained with baseline methods, using only human-labeled data and employing pseudo labels without the incorporation of the corruption pipeline, reinforcing the notion that implementing a corruption pipeline is beneficial for enhancing the performance of artifact detectors.

\subsection{Resampling V.S. Not Resampling}
\begin{figure}[ht]
  \centering
   \includegraphics[width=0.9\linewidth]{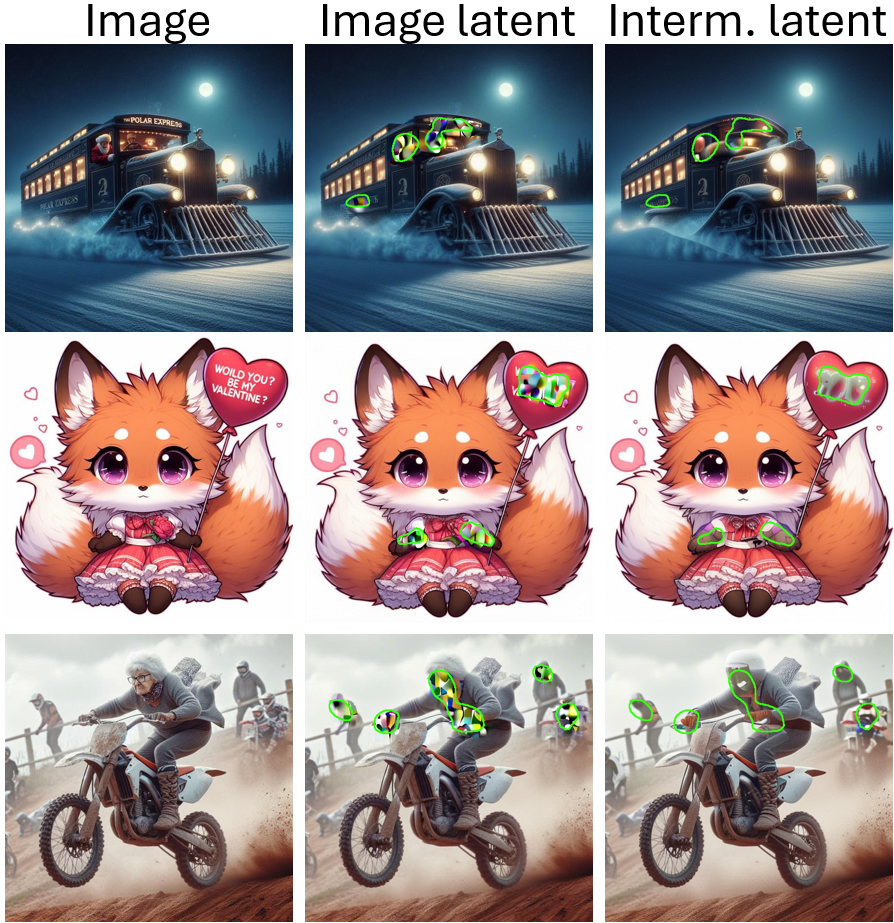}
   \caption{\textbf{Visualization on corruption on image latent \textit{vs.} intermediate latent. and resampling} }
   \label{fig:img_interm_latents}
\end{figure}

Resampling after latent corruption is essential for generating realistic artifacts. In the experiment, we corrupt the latent by substituting the latent with pure Gaussian noise $N(0,1)$ in the selected regions. We compare two settings: (1) direct corruption of the image latent and (2) corruption of an intermediate latent and resampling for 20 steps. In Figure \ref{fig:img_interm_latents}, we observed that direct corruption often results in unnatural, colorful artifacts in the selected region. In contrast, corrupting the intermediate latent and resampling enables the diffusion model to adapt corruptions to the surrounding context, producing man-made artifacts that appear relatively more natural.

To quantify artifact quality, we trained an artifact detector using the corrupted dataset and evaluated it on a human-labeled test set. Table \ref{tab:img_interm_latents} reports that direct corruption yields a mIoU score of 0.22. In contrast, intermediate latent corruption with resampling achieves 12–13 mIoU, yielding a significant difference, underscoring the importance of resampling for realistic artifact synthesis.


\begin{table}[htbp]
\centering
\caption{\textbf{Quantitative analysis on corruption of image latent \textit{vs.} intermediate latent} (20 inversion steps) across model architectures, evaluated on RichHF test set (mIoU score).}
\label{tab:img_interm_latents}
\begin{tabular}{lcc}
\toprule
Corruption Target & Swing-T & ConvNext \\
\midrule
Image Latent & 0.22 & 0.29 \\
Intermediate Latent & \textbf{13.86} & \textbf{12.52} \\
\bottomrule
\end{tabular}
\end{table}

%% file: sec/7_limitation.tex
\section{Limitations}
\label{sec:limitations}
\begin{figure}
    \centering
    \includegraphics[width=0.9\linewidth]{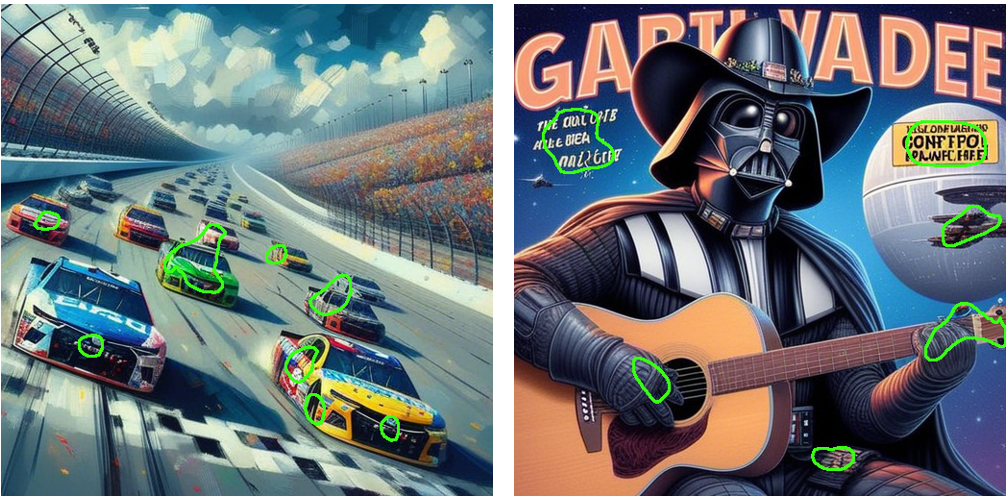}
    \caption{\textbf{Visualization on images that failed to result in artifacts in the selected regions.}}
    \label{fig:failed_imgs}
\end{figure}

Despite the success of the automatic pipeline in generating artifacts to improve the artifact detector, our method isn't guaranteed to generate of artifacts in the selected regions, leading to noisy annotations. For example, in Figure \ref{fig:failed_imgs}, the green marked region in the center of the image of the race car does not result in noticeable artifacts. Also, Darth Vader's right hand lacks detectable artifacts. This limits the performance gains for the artifact detector. The challenge stems from an incomplete understanding of the mechanisms driving artifact generation, with no identified factors guaranteed to trigger artifact generation. Consequently, our approach relies on empirically determined methods that only probabilistically generate artifacts. Future work needs to identify definitive factors linked to artifact generation, enabling a guaranteed pipeline for artifact generation.

%% file: sec/8_conclusion.tex
\section{Conclusion}
In this paper, we introduce an approach to automatically generate pixel-level annotations to form a pixel-level artifact dataset. We proposed an artifact generation pipeline that adds artifacts to specific regions of high-quality synthesized images. We further address technical challenges in keeping artifacts confined and making artifacts look like artifacts in the wild. When trained on our proposed method, the artifact detector can improve as much as 13.2\% compared to baseline approaches. Overall, our work represents an initial step toward building a scalable pipeline for generating pixel-level artifact annotations, aiming to incorporate world knowledge into artifact detection.